%% file: main.tex
\documentclass[runningheads]{llncs}

\usepackage[T1]{fontenc}

\usepackage{graphicx} %
\usepackage{verbatim}
\usepackage{pifont}
\usepackage{ifthen}

\usepackage[table]{xcolor}
\usepackage{amsmath}
\usepackage{amsfonts}
\usepackage{subcaption}

\usepackage{booktabs}   %
\usepackage{multicol}   %
\usepackage{multirow}   %
\usepackage{makecell}   %
\usepackage{caption}

\usepackage{hyperref}

\definecolor{darkgreen}{rgb}{0.0, 0.5, 0.0} %

\hypersetup{
    colorlinks=true,
    linkcolor=blue,
    filecolor=blue,      
    urlcolor=blue,
    citecolor=blue,
}
\begin{document}
\title{TrackOR: Towards Personalized Intelligent Operating Rooms Through Robust Tracking}

\author{
    Tony Danjun Wang\inst{1} \and
    Christian Heiliger\inst{3} \and \\
    Nassir Navab\inst{1,2} \and
    Lennart Bastian\inst{1,2}
} %

\authorrunning{T. Wang et al.}
\institute{Computer Aided Medical Procedures, TU Munich, Germany \and
Munich Center for Machine Learning, Germany \and
Minimally Invasive Surgery, University Hospital of Munich (LMU), Germany
}

\maketitle              %

\begin{abstract}

Providing intelligent support to surgical teams is a key frontier in automated surgical scene understanding, with the long-term goal of improving patient outcomes.
Developing personalized intelligence for all staff members requires maintaining a consistent state of \textit{who} is located \textit{where} for long surgical procedures, which still poses numerous computational challenges.
We propose \textit{TrackOR}, a framework for tackling long-term multi-person tracking and re-identification in the operating room.
\textit{TrackOR} uses 3D geometric signatures to achieve state-of-the-art online tracking performance (+11\% Association Accuracy over the strongest baseline), while also enabling an effective offline recovery process to create analysis-ready trajectories.
Our work shows that by leveraging 3D geometric information, persistent identity tracking becomes attainable, enabling a critical shift towards the more granular, staff-centric analyses required for personalized intelligent systems in the operating room.
This new capability opens up various applications, including our proposed \textit{temporal pathway imprints} that translate raw tracking data into actionable insights for improving team efficiency and safety and ultimately providing personalized support.

\end{abstract}

\input{chapters/1_introduction}

\input{chapters/2_related_works}

\input{chapters/3_method}
\input{chapters/4_experiments_and_results}

\input{chapters/5_conclusion}

\newpage

\bibliographystyle{splncs04}
\bibliography{main.bib}

\newpage

\end{document}

%% file: chapters/1_introduction.tex
\input{figures/teaser_figure}

\section{Introduction}

\noindent
The operating room (OR) is the quintessential high-stakes environment -- small actions can lead to profound consequences.
To improve patient outcomes, Surgical Data Science (SDS) has emerged with the goal of creating a data-driven feedback loop to make the OR safer and more efficient \cite{maier2022surgical}.
Historically, SDS has centered on the surgeon, analyzing instrument handling and technique, often through the viewpoint of endoscopes or laparoscopes \cite{czempiel2022surgical}.
Recognizing that surgery is a complex team endeavor, the field's focus has recently broadened to encompass the entire surgical team's dynamics and collaborative patterns \cite{wang2025beyond}.
Yet this wider perspective still operates at a coarse granularity, typically analyzing workflow at the role level, treating, for instance, the ``circulating nurse'' as an archetype rather than an individual \cite{ozsoy20224d,ozsoy2025mmor}.

We posit that the next generation of surgical domain models will be driven by personalized intelligent systems, which in turn requires a fundamental shift from a role-based to a \textit{staff-centric} understanding of the OR \cite{wang2025beyond}.
This shift moves beyond generic roles by recognizing that each staff member has unique skill levels and develops distinct habits over time, e.g., when adapting to new surgical instruments or team members. 
However, this leap towards personalized intelligence requires new capabilities: generating persistent, long-term trajectories for each staff member, even across extended absences from the OR.

Multi-object tracking (MOT) is a notoriously difficult task with numerous challenges.
Strikingly, the OR represents a confluence of many of these challenges, including severe occlusions and overall crowdedness \cite{bastian2023know}.
These challenges are exacerbated by staff wearing visually indistinct homogeneous attire, which has rendered prior OR tracking methods fundamentally identity-agnostic.
While these methods can handle the simple, short-term task of associating an individual from one frame to the next, they inevitably fail when faced with the `revolving door\footnote{Refers to, e.g., the circulating nurse shuttling between ORs}` reality of surgical procedures, confining their utility to uninterrupted video segments and making longitudinal analysis challenging.

To overcome this, a robust global re-identification (ReID) capability is not just beneficial but essential.
While many tracking methods incorporate appearance-based ReID \cite{ciaparrone2020deep}, such methods are bound to fail when confronted with the visual homogeneity of the OR \cite{wang2025beyond}.
A more powerful paradigm is needed, where identity is derived from a robust, view-invariant signature that is decoupled from confounding visual and textural cues.
The effectiveness of such a paradigm, however, depends largely on the choice of its underlying data representation, for example, 2D appearance, 3D pose, or 3D geometry (see \autoref{fig:teaser}).
In this paper, we investigate these options and propose a framework centered on 3D geometric signatures as a robust solution.
Our approach leverages point cloud representations for online association, also enabling an offline process to correct tracking errors and merge fragmented tracklets, reconstructing complete, persistent journeys of each staff member.
Our \textbf{contributions} are summarized as follows:

\begin{itemize}
    \item We propose \textit{TrackOR}, a novel end-to-end tracking framework that overcomes the limitations of appearance-based methods by integrating a robust, view-invariant ReID signature for persistent, long-term identity tracking.
    \item We introduce a method that combines 3D pose and ReID for online tracking and leverages ReID alone for offline global trajectory recovery, showing a unique synergy for solving the tracking problem in the OR.
    \item We propose \textit{temporal pathway imprints} and demonstrate how our framework unlocks long-term workflow and pathway analysis with these imprints.
\end{itemize}

%% file: figures/teaser_figure.tex
\begin{figure}[t!h]
    \centering
    \includegraphics[width=\columnwidth]{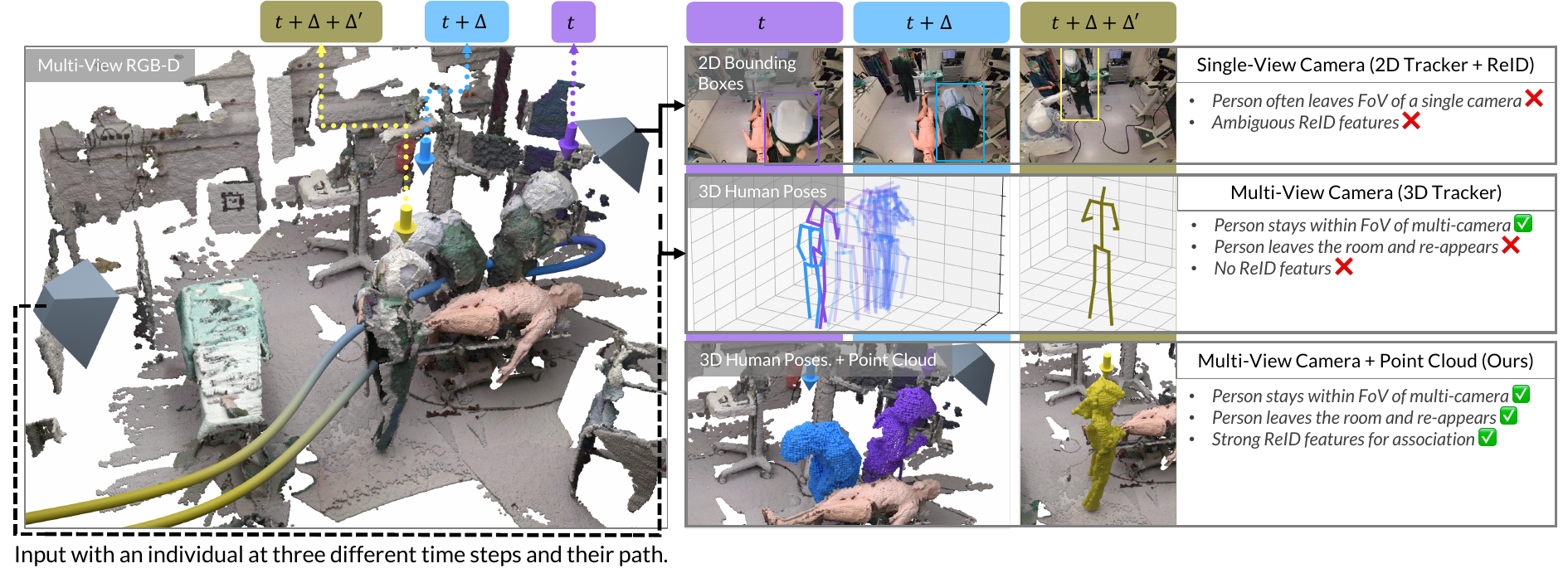}
    \vspace{-.6cm}
    \caption{
    We introduce \textit{TrackOR}, a framework for long-term multi-person tracking in the OR. Unlike conventional 2D trackers that struggle with field-of-view (FoV) limits and appearance-based ReID, or standard 3D trackers that lack features for re-identification after prolonged absences, \textit{TrackOR} uses 3D geometric signatures to maintain persistent identity even when staff leave and re-enter the room.
    }
    \label{fig:teaser}
    \vspace{-0.7cm}
\end{figure}

%% file: chapters/2_related_works.tex
\section{Related Works}

Multi-Object Tracking (MOT) represents a formidable challenge in the operating room (OR). 
While standard trackers typically combine a motion model, like the Kalman filter \cite{kalman}, for short-term prediction \cite{cao2023observation,zhang2022bytetrack} with an appearance based Re-Identification (ReID) module to handle longer occlusions \cite{du2023strongsort,aharon2022bot,stanojevic2024boosttrack++,maggiolino2023deep}, these approaches are confounded by the complex dynamics and visually homogeneous OR environment. 
Consequently, prior OR-specific approaches, whether they reconstruct and associate in 3D directly ("3D-first") \cite{belagiannis2016parsing,ozsoy20224d} or track in 2D before aggregating ("2D-first") \cite{huMultiCameraMultiPersonTracking2022a}, have largely been identity-agnostic, limiting their utility to short-term analysis. 
While recent work has shown the promise of non-texture-based signatures for the isolated ReID task \cite{wang2025beyond}, a complete framework that effectively integrates this principle for robust, long-term tracking has remained elusive.
Our work bridges the gap between the isolated Re-Identification and robust end-to-end tracking; this comprehensive framework leverages non-texture-based signatures to recover the global, long-term trajectories of OR staff.

%% file: chapters/3_method.tex
\input{figures/main_figure}
\section{Unifying Short- and Long-term Tracking}

Given a video sequence of $T$ frames, each frame $t$ contains a set of detections $D_t = \{d_t^1, d_t^2, \dots, d_t^{N_t}\}$.
Our objective is to solve a multi-object tracking problem with long-term re-identification.
The goal is to partition the detections from all frames into a set of trajectories $\mathcal{T} = \{\tau_1, \tau_2, ..., \tau_M\}$, where $M$ is the total number of unique individuals observed.

In the OR, however, the tracking problem is compounded by staff frequently leaving and re-entering the scene for extended periods.
These long-term absences cause simple, continuous trajectory models to be ineffective.

We first define a \textit{tracklet} $\text{trk}_m^a$ as a single, continuous period of visibility for an individual $m$.
It is formally defined as a time-ordered sequence of states: 

\[
\text{trk}_m^a = \{s_t^m\}_{t=b_m^a}^{e_m^a},
\]

where $s_t^m$ is the state of individual $m$ at frame $t$ (e.g., a bounding box), and the indices $b_m^a$ and $e_m^a$ denote the beginning and ending frames of that specific appearance.
Building on this, we define a \textit{trajectory} $\tau_m$ as the complete history of an individual, represented by a collection of their tracklets:

\[
\tau_m = \{ \text{trk}_m^1, \text{trk}_m^2, ..., \text{trk}_m^{A_m}\},
\]

where $A_m$ is the number of distinct appearances for individual $m$.
This formulation explicitly models the periods of absence between tracklets, making it well-suited for the tracking problem in the OR.
This formulation encompasses both short-term and long-term absences between different surgical procedures.

\subsection{TrackOR}

In \autoref{fig:main}, we illustrate the overall pipeline of our proposed method, TrackOR.
TrackOR follows the ``tracking by detection'' paradigm \cite{ciaparrone2020deep}, consisting of detection, feature extraction, and association (\autoref{fig:main} top).
Moreover, with our framework, we can further leverage extracted ReID features to perform a straightforward and effective offline global trajectory recovery (\autoref{fig:main} bottom).

\noindent
\textbf{Detection.}
Following prior `3D-first' approaches \cite{belagiannis2016parsing,ozsoy20224d,liu2024human}, we directly detect 3D human poses from the multi-view RGB data.
Specifically, we adopt VoxelPose \cite{tu2020voxelpose}, which aggregates heatmap predictions from each 2D view into a unified 3D voxel volume.
This volume is then processed by a 3D CNN that predicts the root location of each person.
For each predicted root location, another 3D CNN then regresses the final, detailed 3D poses.
For a complete description of this process, we refer the reader to the original publication \cite{tu2020voxelpose}.

\noindent
\textbf{Feature Extraction.}
We obtain the person ReID features $\mathcal{F}_t^{i}$ from the segmented 3D point cloud of each 3D human pose detection.
Specifically, we first segment the 3D point cloud of the entire scene to obtain 3D point clouds of each human and associate each 3D human pose with their respective 3D human point cloud \cite{bastian2023segmentor}.
Subsequently, each 3D human point cloud is projected into 8 virtual camera viewpoints, positioned around each object in an equidistant circular arrangement, to generate a set of 2D depth maps that are further processed with a ReID network \cite{wang2025beyond} to obtain our final ReID feature vectors $\mathcal{F}_t^{i} \in \mathbb{R}^{8 \times C}$, where $C$ is the feature dimension.

\noindent
\textbf{Association.}
At this point, for each frame $t$, we have a set of 3D human pose detections $D_t$ and a set of active trajectories  $\mathcal{T}_{t-1}$ from the previous frame.
Each detection $d_t^i$ and trajectory $\tau_m$ is represented by a ReID feature vector and a 3D bounding box.
To associate new detections with existing trajectories, we compute a cost matrix $\mathbf{C}$ where each entry represents the dissimilarity between a detection and a trajectory. 
We define the cost as a weighted sum of the shape cost, based on the cosine dissimilarity of the ReID features, and a spatial cost, based on the 3D Generalized IoU (GIoU) \cite{Rezatofighi_2018_CVPR} between bounding boxes. 
With the cost matrix defined, we employ a linear assignment strategy using the Hungarian algorithm \cite{kuhn1955hungarian} to find the optimal matching.
Associations with a final cost higher than a predefined threshold $\gamma$ are discarded.
Matched detections update their respective trajectory states, unmatched detections initialize new tracks, and unmatched trajectories are marked as ``lost''.

\noindent
\textbf{Global Trajectory Recovery.}
For downstream tasks, we perform a final offline recovery step to correct fragmentation and identity switches from the online stage (\autoref{fig:main} bottom). 
To do this, we first apply temporal max-pooling to aggregate each tracklet's sequence of feature vectors $\mathbb{R}^{l \times 8 \times C}$ into a single representative descriptor per view.
An SVM-Gallery \cite{wang2025beyond,layne2017dataset} then assigns an identity to this descriptor via a majority vote over its 8 view-specific feature vectors. 
Finally, all tracklets assigned the same identity are grouped to reconstruct the complete, global trajectory for each person.

\noindent
\textbf{Temporal Pathway Imprint.}
To extract insights from the obtained global trajectories, we propose \textit{temporal pathway imprints}.
We achieve this by projecting the root positions onto the X-Y plane of the OR and viewing the result from a bird's-eye view.
Adding regions of interest, such as sterility zones, can further enhance the context.
Analagous to Wang et al.'s \cite{wang2025beyond} \textit{3D activity imprints} which depict the duration of time at each spatial location, we model an individual's temporal pathway through the OR as a function of time.

%% file: figures/main_figure.tex
\begin{figure}[t!]
    \centering
    \includegraphics[width=\columnwidth]{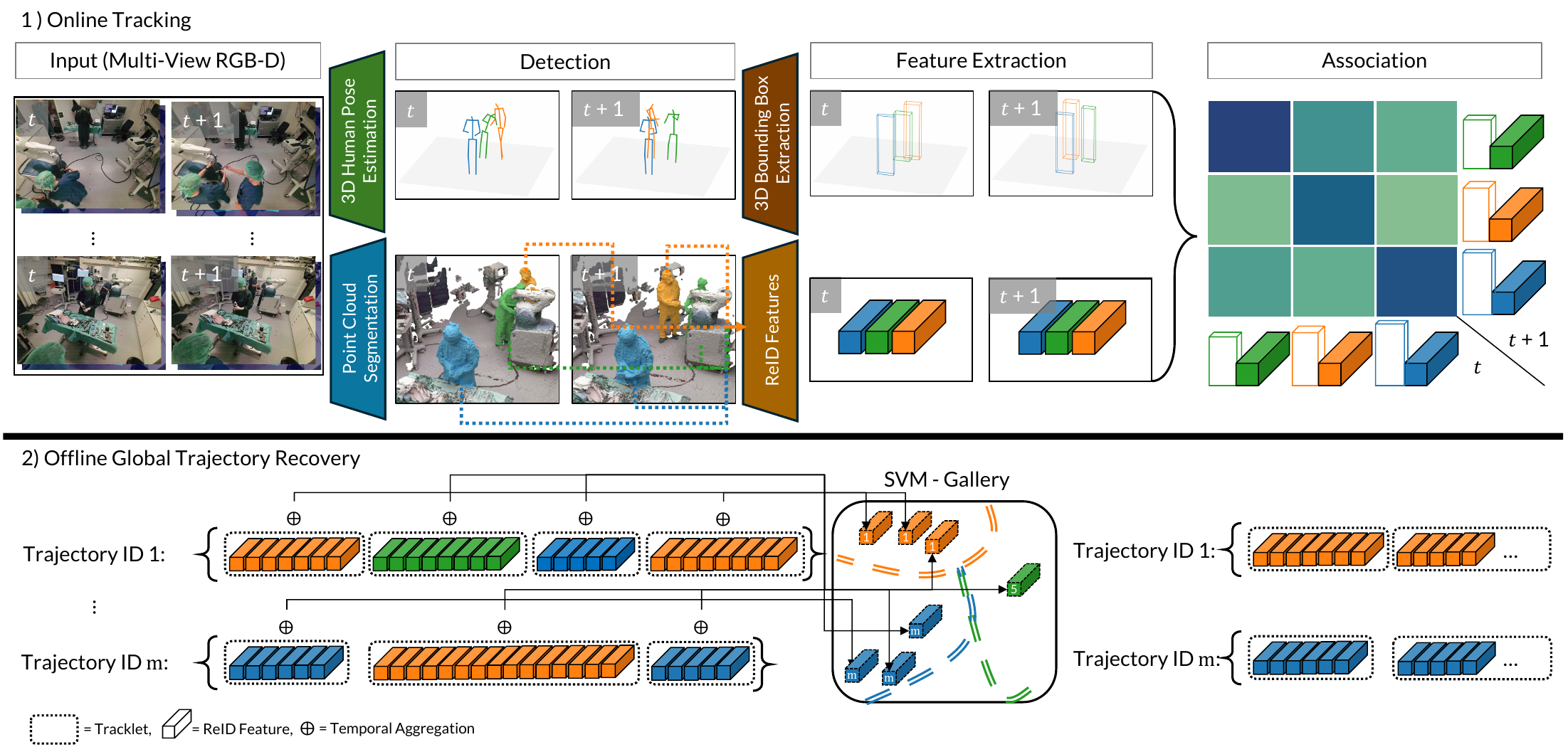}
    \caption{
    Overview of our online (top) and offline trajectory recovery (bottom).
    Online, we use multi-view RGB-D frames to extract spatial and ReID features from 3D poses and point clouds, creating a cost matrix for data association. Offline, we classify the resulting tracklets using an SVM-Gallery \cite{wang2025beyond} and reconstruct each person's global trajectory based on this classification.
    }
    \label{fig:main}
\end{figure}

%% file: chapters/4_experiments_and_results.tex
\section{Experiments and Results}
\input{tables/tracking}

\noindent 
\textbf{Datasets.}
Due to privacy concerns, publicly available datasets of surgical procedures captured from ceiling-mounted cameras are generally scarce \cite{bastian2023disguisor}. 
While previous datasets such as MVOR \cite{srivastav2018mvor} and 4D-OR \cite{ozsoy20224d} exist, they are not suitable for developing robust, long-term tracking solutions. 
MVOR \cite{srivastav2018mvor} lacks the necessary temporal continuity and identity annotations, while 4D-OR \cite{ozsoy20224d} is overly simplified due to minimal movement and obstructions, allowing near-perfect tracking with a simple nearest-neighbor association \cite{wang2025beyond}. 
As such, we perform our experiments on the recently introduced MM-OR dataset \cite{ozsoy2025mmor}, which, in contrast, exhibits tracking challenges, including frequent occlusions, multiple clinicians working in tight spaces, and homogenous attire.
For our experiments, we use the multi-view RGB-D data from the ceiling-mounted cameras, which provide three annotated RGB views and a corresponding 3D point cloud for each frame.
In our study, we use all takes in MM-OR that provide segmentation labels. From the resulting 20 takes (totaling 23,442 frames), we partition 13 takes (62\%) into a train set, 2 takes (16\%) into a validation set, and 5 takes (22\%) into a test set.
In MM-OR, each take corresponds to a single surgery.
We consulted the dataset's authors to obtain identity mappings.

\noindent
\textbf{Metrics.} 
To evaluate the broad range of tracking methods (2D and 3D) consistently, we assess all methods in the 2D image space.
Accordingly, maintaining consistent IDs across different camera views is not required by our evaluation protocol.
For 3D methods, we project 3D pose detections into each 2D image plane to generate bounding boxes.
We follow standard multi-object tracking benchmarks and report the Higher Order Tracking Accuracy (HOTA) \cite{luiten2021hota} with its sub-components, as well as the classic CLEAR \cite{bernardin2008evaluating}, Identity \cite{ristani2016performance}, and Counting metrics.
To account for the inherent geometric discrepancies between bounding boxes derived from 3D poses and the 2D ground truth derived from silhouettes, we adjust the HOTA $\alpha$-range to $0.05-0.5$.

\input{figures/qualitative_figure}

\subsection{Implementation Details}

\noindent
\textbf{Detection Backbones.}
To ensure a fair comparison, all evaluated methods use detections from the same backbone.
For 2D tracking baselines, we provide detections from a YOLOv10-B \cite{wang2024yolov10} model fine-tuned on our training set.
For all 3D-based methods (including ours), we use 3D human pose estimations as detection.
Since MM-OR lacks 3D annotations, we train a pose estimation network using the self-supervised approach proposed in \cite{Srivastav_2024_CVPR}.
3D point cloud segmentations are obtained by projecting ground truth segmentation masks into 3D.

\noindent
\textbf{Re-identification Modules.} 
The appearance-based 2D trackers utilize a standard feature extractor \cite{luo2019bag} fine-tuned on cropped images of individuals from our training split.
In contrast, our method's 3D ReID module uses a ResNet-9 backbone \cite{wang2025beyond} to extract view-invariant features from person-specific point clouds, likewise trained on the identity labels of our training set.

\noindent
\textbf{Evaluated Trackers.}
We compare our framework against a comprehensive set of baselines.
The 2D trackers include OCSort \cite{cao2023observation} and ByteTrack \cite{zhang2022bytetrack} (without ReID), as well as the ReID-based methods BoostTrack++ \cite{stanojevic2024boosttrack++}, DeepOCSort \cite{maggiolino2023deep}, StrongSort \cite{du2023strongsort}, and BoTSort \cite{aharon2022bot}.
Our 3D baselines implement the tracking paradigms proposed by Belagiannis et al. \cite{belagiannis2016parsing,KSP_tracker}, Ozsoy et al. \cite{ozsoy20224d}, and Liu et al. \cite{liu2024human}. For all methods, tracking hyperparameters are tuned on the validation set.

\subsection{Results}
\input{figures/downstream_task_figure}

\noindent
\textbf{Tracking Performance.}
The quantitative results in \autoref{tab:benchmark} show that the online results of TrackOR (Ours) score the highest overall HOTA \cite{luiten2021hota} of 82.2\%.
This performance is driven by the Association Accuracy (AssA) of 82.3\%, which is over 11\% higher than the best baseline (BoT Sort \cite{aharon2022bot}).
This result confirms the superiority of our geometric ReID feature for maintaining correct identities through the challenging occlusions and visual homogeneity of the OR.
Conversely, the results reveal that all 3D-based methods exhibit a lower detection accuracy (DetA) compared to the 2D baselines.
This is an expected trade-off, as our 3D pose estimation backbone is trained using a self-supervised framework, compared to the 2D detector fine-tuned on ground-truth bounding boxes.
Consequently, older metrics that are heavily biased towards detection, like MOTA \cite{bernardin2008evaluating}, naturally favor the 2D approaches.
This trade-off also extends to computational cost, as processing 3D data results in a lower framerate (FPS) for TrackOR compared to the faster, single-modality 2D trackers (measured on a single RTX 2080 Ti).

\autoref{fig:qualitative} illustrates qualitative results of BoT Sort (with ReID) \cite{aharon2022bot} and TrackOR.
The sequence highlights a common failure mode where 2D trackers lose a person during occlusion and subsequently switch their ID due to visual similarity. 
Our method, operating in 3D, successfully navigates both the occlusion and the re-identification challenge, maintaining the correct identities throughout.

\noindent
\textbf{Downstream Task: Temporal Pathway Imprints.}
As a downstream application of our tracker, \autoref{fig:pathways} shows two \textit{temporal pathway imprints}  from the same robot technician across two different surgeries.
The imprints reveal that while the technician's primary workspace was the MPS station in both procedures, they interacted with the robot twice in Surgery 1 compared to only once in Surgery 2.
More critically, the pathway in Surgery 2 captures the non-sterile technician coming into close proximity with the sterile patient table. 
Ultimately, these imprints demonstrate the potential to move towards a data-driven science of the OR, enabling automated workflow analysis, objective safety monitoring, and personalized feedback for the entire surgical team.

%% file: tables/tracking.tex
\begin{table*}[t!]
    \centering
    \setlength{\tabcolsep}{3pt}
    \caption{
    Quantitative comparison of TrackOR (Ours) against 2D and 3D tracking baselines on the MM-OR test set. 
    Bold indicates the best performance per metric. $\uparrow$ Higher is better, $\downarrow$ lower is better, $\dag$ denotes offline methods.}
    \resizebox{\linewidth}{!}{%
    \begin{tabular}{@{}lc c ccc c cc c ccc c cc c r@{}}
        \toprule
        \multicolumn{2}{c}{Model} & &
        \multicolumn{3}{c}{$\text{HOTA}_{(\alpha=.05-.5)}$ \cite{luiten2021hota}} & &
        \multicolumn{2}{c}{Identity\cite{ristani2016performance}} & &
        \multicolumn{3}{c}{CLEAR \cite{bernardin2008evaluating}} & &
        \multicolumn{2}{c}{Count} & &
        \multicolumn{1}{c}{Speed} \\
        \cmidrule{1-2}
        \cmidrule{4-6}
        \cmidrule{8-9}
        \cmidrule{11-13}
        \cmidrule{15-16}
        \cmidrule{18-18}
        Tracker &
        ReID & &
        HOTA$\uparrow$ &
        AssA$\uparrow$ &
        DetA$\uparrow$ & &
        IDF1$\uparrow$ &
        IDSW$\downarrow$ & &
        MOTA$\uparrow$ &
        FP$\downarrow$ &
        FN$\downarrow$ & &
        \% \#Dets & %
        \% \#IDs & & %
        FPS$\uparrow$ \\
        \midrule \midrule
        \multicolumn{18}{c}{\textbf{2D Bounding Box Tracker, using \cite{wang2024yolov10} as detections.}} \\
        \midrule
        OC-Sort \cite{cao2023observation} & \ding{55}           && 49.660 & 27.054 & 91.158 && 40.089 & 566 && 79.575 & 849 & \textbf{566} && 102.92 & 362.26 && 850 \\
        ByteTrack \cite{zhang2022bytetrack} & \ding{55}         && 58.430 & 37.451 & 91.163 && 52.705 & 312 && 75.946 & 950 & 1071 && 98.75 & 141.50 && 997 \\
        Strong Sort \cite{du2023strongsort} & RGB               && 57.965 & 36.783 & 91.347 && 43.378 & 377 && 72.729 & 679 & 744 && \textbf{99.33} & 164.15 && 20 \\
        Boost Track \cite{stanojevic2024boosttrack++} & RGB     && 54.848 & 33.013 & 91.130 && 42.989 & 511 && 77.420 & 622 & 1057 && 95.52 & 1,037.74 && 60 \\
        Deep OC-Sort \cite{maggiolino2023deep} & RGB            && 78.359 & 66.348 & \textbf{92.545} && 73.007 & 200 && \textbf{85.256} & \textbf{437} & 793 && 97.36 & 239.62 && 34 \\
        BoT Sort \cite{aharon2022bot} & RGB                     && 80.825 & 71.309 & 91.612 && 74.686 & 266 && 78.936 & 837 & 940 && 98.94 & 139.62 && 32 \\
        \midrule \midrule
        \multicolumn{18}{c}{\textbf{3D Human Pose Tracker, using \cite{Srivastav_2024_CVPR} as detections.}} \\
        \midrule
        KSP Tracker$^\dag$ \cite{belagiannis2016parsing} & \ding{55}       && 54.037 & 36.086 & 80.918 && 46.369 & 462 && 51.768 & 1566 & 2650 && 88.82 & 158.49 && 115 \\
        Nearest-Neighbor \cite{ozsoy20224d} & \ding{55}             && 73.366 & 65.813 & 81.787 && 66.496 & 86 && 55.686 & 1564 & 2648 && 88.82 & 113.21 && 2365 \\
        Kalman Filter \cite{liu2024human} & \ding{55}               && 71.047 & 62.441 & 80.840 && 63.317 & \textbf{80} && 54.686 & 1543 & 2772 && 88.82 & \textbf{101.88} && 1121 \\
        TrackOR (Ours)    & Depth                                   && \textbf{82.216} & \textbf{82.300} & 83.685 && \textbf{76.362} & 125 && 55.284 & 1564 & 2648 && 88.82 & 130.19 && 17 \\
        \bottomrule
    \end{tabular}
    }
    \label{tab:benchmark}
\end{table*}

%% file: figures/qualitative_figure.tex
\begin{figure}[t!]
    \centering
    \includegraphics[width=\columnwidth]{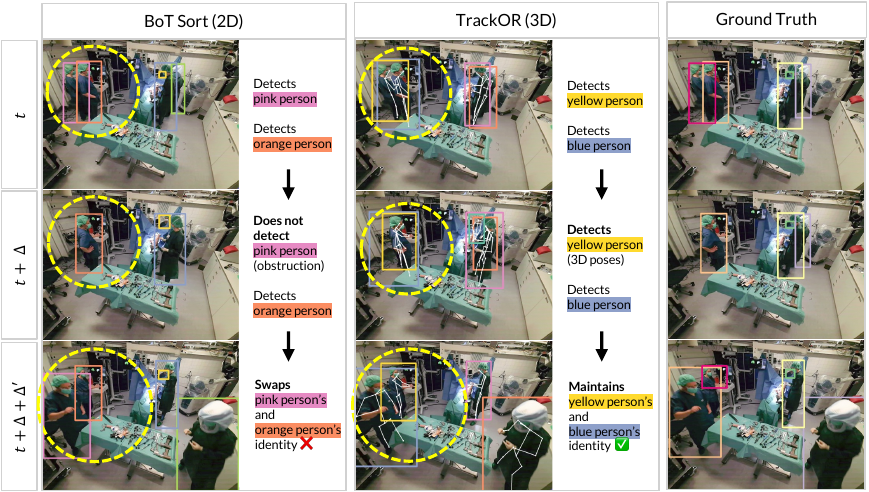}
    \caption{Qualitative results of BoT-Sort \cite{aharon2022bot} and TrackOR (Ours).
    The bounding box colors reflect the predicted identity.
    }
    \label{fig:qualitative}
\end{figure}

%% file: figures/downstream_task_figure.tex
\begin{figure}[t!h]
    \centering
    \includegraphics[width=\columnwidth]{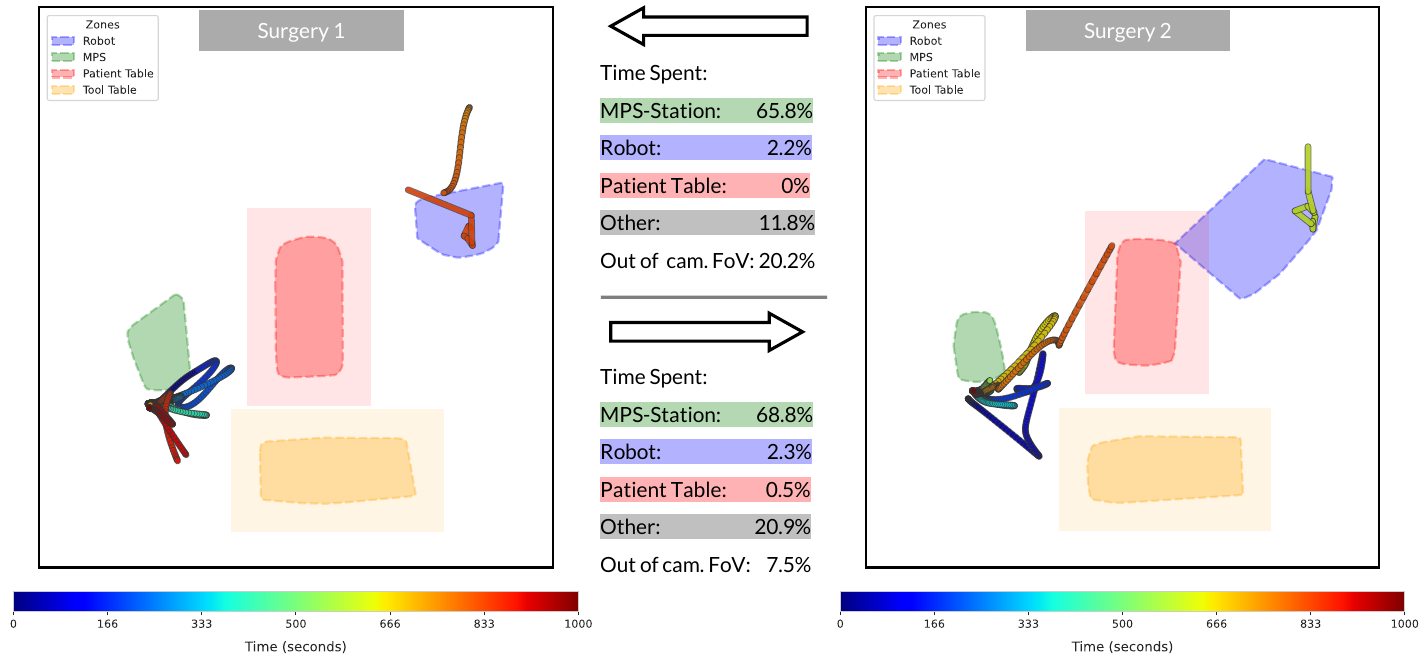}
    \caption{\textit{Temporal pathway imprints} of the robot technician of two different surgeries. We plot the first 1,000s of each surgery. The extended transparent borders delineate the 12-inch border around the sterile fields.
    }
    \label{fig:pathways}
\end{figure}

%% file: chapters/5_conclusion.tex
\section{Conclusion}

We presented TrackOR, a novel framework for robust, long-term multi-person tracking and re-identification in the operating room. 
Prior methods often fail to maintain identity through the prolonged absences of staff common in surgery, a challenge exacerbated by visually similar attire that confounds appearance-based ReID. 
Our approach addresses these challenges by deriving identity from robust 3D geometric signatures, which enables fine-grained, person-centric workflow analyses, as demonstrated by our \textit{temporal pathway imprints}.
We will make all codes public as we believe this work is an important step towards the next generation of personalized intelligent systems capable of providing tailored support to the entire surgical team.

\begin{credits}
\subsubsection{\ackname}
This work was partly supported by the state of Bavaria through Bayerische Forschungsstiftung (BFS) under Grant AZ-1592-23-ForNeRo and the German Federal Ministry for Economic Affairs and Climate Action (BMWK) through the Central Innovation Programme for SMEs (ZIM) under Grant KK 5389102BA3.

\subsubsection{\discintname}
The authors have no competing interests to declare that are
relevant to the content of this article.
\end{credits}